%% file: main.tex
\documentclass[11pt]{article}
\usepackage[margin=1in]{geometry}
\usepackage{amsmath,amssymb,amsthm,mathtools,bm}
\PassOptionsToPackage{protrusion=false}{microtype}
\usepackage{graphicx,booktabs,array,microtype,xcolor}
\usepackage[numbers,sort&compress]{natbib}
\usepackage[colorlinks=true,linkcolor=blue!40!black,citecolor=blue!40!black,urlcolor=blue!40!black]{hyperref}
\usepackage{enumitem,etoolbox}
\makeatletter
\patchcmd{\@maketitle}{\vskip 2em}{\vskip -0.5em}{}{}
\patchcmd{\@maketitle}{\vskip 1.5em}{\vskip 1em}{}{}
\makeatother
\setlist{nosep,leftmargin=*}
\newcolumntype{P}[1]{>{\raggedright\arraybackslash}p{#1}}
\newtheorem{theorem}{Theorem}[section]
\newtheorem{proposition}[theorem]{Proposition}

\theoremstyle{definition}
\theoremstyle{remark}
\newcommand{\E}{\mathbb E}
\newcommand{\KL}{D_{\mathrm{KL}}}
\newcommand{\R}{\mathcal R}

\newcommand{\cA}{\mathcal A}
\newcommand{\tr}{\operatorname{tr}}

\title{A Recipe for Physical Action Model Training}
\author{Shaojiang Wang$^{1}$ \qquad Geer Yang$^{2,1,3}$\\[5pt]
\small $^1$Institute of AI for Industries, Chinese Academy of Sciences, Nanjing, China\\
\small $^2$Institute of Computing Technology, Chinese Academy of Sciences, Beijing, China\\
\small $^3$University of Chinese Academy of Sciences, Beijing, China}
\date{}
\hypersetup{pdftitle={A Recipe for Physical Action Model Training},pdfauthor={Shaojiang Wang and Geer Yang}}
\begin{document}
\maketitle
\begin{abstract}
Choosing robot model capacity and demonstration budgets also requires decisions about data composition, representation, optimization, and deployment. We connect these decisions through a common conditional-distribution target. Nested learning oracles give exact risk accounting; local information geometry, conditional mutual information, and empirical convergence provide conditional training prescriptions. Known-target numerical checks expose capacity--data coupling, representation-induced mixture mismatch, and an example where deployment KL increases while negative log-likelihood decreases. An analysis of 206 public PushT trajectories evaluates fixed-feature information-design criteria without fitting a policy. Under asymmetric evaluation targets, the selected allocations reduce this criterion by 4.5--6.2\% relative to target matching; balanced targets show no benefit. These results clarify which quantities can support training recommendations and when an apparently favorable predictive metric has a different interpretation.
\end{abstract}

\section{Introduction}
Robot behavior cloning couples several resource decisions. Equally sized datasets can provide different information about relevant actions; equally sized encoders can preserve different information; and converged optimization can leave a poorly matched policy. A useful framework must identify which intervention deserves the next unit of data or compute.

\begin{figure}[!t]
\centering
\includegraphics[width=\linewidth]{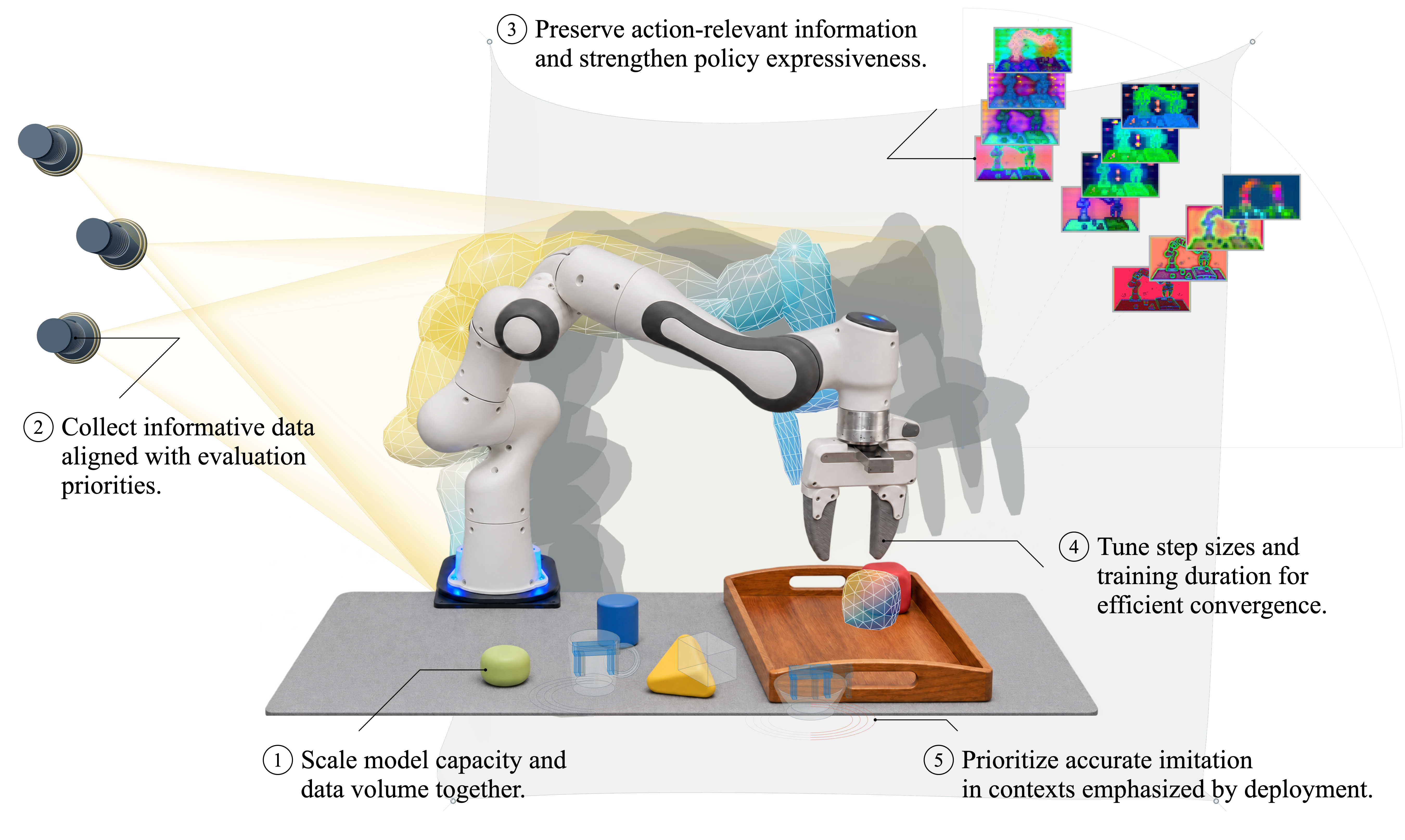}
\caption{Five training decisions under a common behavior-cloning objective. The scene is a generatively edited illustration; meshes and projections depict conceptual mappings. The twelve panels show pretrained ResNet-50, CLIP, and DINOv2 features at four depths, visualized by independent RGB PCA. Colors describe within-layer variation rather than shared semantic labels.}
\label{fig:framework}
\end{figure}

We compare these decisions under one conditional expert-action target. Histories may contain vision, language, proprioception, and past interaction. Holding this target fixed connects capacity, acquisition, representation, optimization, and deployment while keeping their assumptions distinct (Figure~\ref{fig:framework}).

\paragraph{Related work.}
Prediction-loss scaling depends on data, compute, architecture, and tokenization \citep{kaplan2020,hoffmann2022,pearce2025}. Robot studies examine demonstration diversity \citep{lin2024scaling} and distributionally robust mixtures \citep{hejna2025}; our acquisition criterion instead specifies a fixed evaluation target. Action-token and continuous-action systems further motivate a common prediction protocol \citep{brohan2023,kim2024,black2024}. The analysis connects proper scoring rules, information theory, optimal design, likelihood asymptotics, and conditional optimization guarantees \citep{gneiting2007,cover2006,pukelsheim2006,vaart1998,karimi2016}.

\paragraph{Contribution.}
We formulate five conditional training relations with explicit interfaces and distinguish learning risk from approximation and empirical convergence. Complete proofs, known-target numerical checks, and a public-trajectory design study show how their prescriptions depend on assumptions. The evidence concerns analytical quantities and state geometry, not trained-policy performance.

\clearpage

\section{Prediction target and risk accounting}\label{sec:target}
Let $H$ be the available history and $A$ an action under a fixed measurement and decision protocol. An unknown expert $\xi$ has conditional law $p_\xi(a\mid h)$ and reference history distribution $d_{E,\xi}$. The imitation risk is
\begin{equation}\label{eq:target}
\R_{E,\xi}(\pi)=\E_{H\sim d_{E,\xi}}\KL\!\left(p_\xi(\cdot\mid H)\|\pi(\cdot\mid H)\right).
\end{equation}
All logarithms are natural. The action units, chunk duration, decoder, and conditioning variables are common across comparisons. Continuous distributions require a common dominating measure and finite KL; a deterministic continuous target is not interchangeable with a smooth density.

For a fixed target with finite entropy and log loss, expected negative log-likelihood (NLL) satisfies $C_E(\pi)=\mathsf H_E(A\mid H)+\R_E(\pi)$. Consequently, differences between policies on the same target compare imitation risks without knowing expert entropy. Absolute NLL is a predictive metric, not an absolute KL estimate. In closed-loop control the visited histories also depend on the policy, so prediction and task performance remain distinct \citep{ross2011}.

\paragraph{Unknown experts and common information.}
Fix a prior $\rho$ over experts before optimization. A design $q\in\mathcal Q$ specifies the acquisition distribution, $g$ a representation, and $\cA\in\mathcal A_{N,g}$ a learner with an output-capacity budget $N$. With data $S_D\sim P_{\xi,q}^{D}$ and learner randomness $U$, define
\begin{equation}\label{eq:bayesfunctional}
\mathcal B(q,g,\cA)=\E_{\xi\sim\rho,S_D,U}\R_{E,\xi}\bigl(\cA(S_D;U)\bigr).
\end{equation}
The baseline treats $D$ as independent observations. Trajectory sampling is addressed explicitly in the public-data study. Pretrained information is either common side information or part of the resource budget. Taking an oracle separately for a known expert would permit hard-coding that expert and would not define a learning limit.

For fixed $N,D$, let
\begin{align}\label{eq:oracles}
L^{(0)}&=\inf_{q',g',\cA'\in\mathcal A_{N,g'}}\mathcal B(q',g',\cA'), &
L^{(1)}&=\inf_{g',\cA'\in\mathcal A_{N,g'}}\mathcal B(q,g',\cA'),\\
L^{(2)}&=\inf_{\cA'\in\mathcal A_{N,g}}\mathcal B(q,g,\cA'), &
L^{(3)}&=\mathcal B(q,g,\cA).\nonumber
\end{align}
For the same expert, data, and learner aggregation, let $L_{\rm dep}=\E_{\xi,S_D,U}\R_{{\rm dep},\xi}(\cA(S_D;U))$. Define $\Delta_{\rm data}=L^{(1)}-L^{(0)}$, $\Delta_{\rm repr}=L^{(2)}-L^{(1)}$, $\Delta_{\rm alg}=L^{(3)}-L^{(2)}$, and $\Delta_{\rm shift}=L_{\rm dep}-L^{(3)}$. For finite risks and nested feasible classes,
\begin{equation}\label{eq:master}
\boxed{L_{\rm dep}=L^{(0)}+\Delta_{\rm data}+\Delta_{\rm repr}+\Delta_{\rm alg}+\Delta_{\rm shift}.}
\end{equation}
The first three gaps are nonnegative because restrictions cannot decrease an infimum; the identity follows by telescoping. The shift term is signed. The order is an accounting convention: interactions can change each gap, and $\Delta_{\rm alg}$ includes statistical and regularization effects as well as optimization.

\section{Five conditional training relations}
\subsection{Capacity and data}\label{sec:counts}
Nested policy families and admissible sample subsets imply that $L^{(0)}$ is nonincreasing in either budget. They do not determine an exponent. A candidate response surface is
\begin{equation}\label{eq:candidate}
L^{(0)}(N,D)\approx L_\infty+aN^{-\alpha}+cD^{-\beta},\qquad a,c,\alpha,\beta>0.
\end{equation}
If this form is supported over the relevant range and compute obeys $F=\kappa ND$, an interior allocation satisfies $\alpha aN^{-\alpha}=\beta cD^{-\beta}$. Equalizing the two unweighted contributions instead gives a crossover, not generally the compute optimum. Both prescriptions depend on the response and cost models.

An exactly solvable calibration case exposes a different capacity--data coupling. Let expert action means be independent $\theta_j\sim\mathcal N(0,\tau_j^2)$, with $\tau_1^2\ge\cdots\ge\tau_m^2>0$, and observe $D$ independent vectors $A_r\sim\mathcal N_m(\theta,\sigma^2I)$. Policies learn the first $N$ means and fix the rest to zero. The optimal Bayes risk is
\begin{equation}\label{eq:exactgaussian}
L_G^\star(N,D)=\frac{1}{2\sigma^2}\left[\sum_{j>N}\tau_j^2+\sum_{j=1}^N\frac{\tau_j^2\sigma^2}{\sigma^2+D\tau_j^2}\right].
\end{equation}
Here $N$ counts trainable coordinates and $D$ counts vector observations. The prior spectrum determines the approximation curve, while the estimation term couples both budgets. This supplies a known target for numerical verification, not a fitted neural scaling law. A complementary finite-description bound is given in Appendix~\ref{app:finite}.

\subsection{Target-weighted data allocation}\label{sec:data}
Fix a regular, identifiable, correctly specified conditional model $p_\theta(A\mid Z)$ of fixed dimension, where $Z$ is the actual model input. Strata change its input distribution while sharing the same conditional law. At the true parameter, let $F_i$ be the Fisher information of stratum $i$, $F(q)=\sum_iq_iF_i\succ0$, and $G_E$ the evaluation-risk Hessian. Under the moment and remainder assumptions stated in Appendix~\ref{app:proofs},
\begin{equation}\label{eq:fisher}
\E\R_E(p_{\hat\theta_D})=\frac{V(q)}{2D}+o(D^{-1}),\qquad V(q)=\tr\!\left[G_EF(q)^{-1}\right].
\end{equation}
For fixed admissible $q$ and $q^\star$, the leading risk difference is $[V(q)-V(q^\star)]/(2D)$. This local comparison holds the model fixed; it is not the globally reoptimized oracle gap in Eq.~\eqref{eq:master}.

For counts $n_i$ and $\mathcal I=\sum_in_iF_i$, the marginal value of one sample is
\begin{equation}\label{eq:marginal}
v_i=-\frac{\partial}{\partial n_i}\frac12\tr(G_E\mathcal I^{-1})
=\frac12\tr(G_E\mathcal I^{-1}F_i\mathcal I^{-1}).
\end{equation}
At a continuous optimum, active strata equalize $v_i/c_i$, where $c_i$ is acquisition cost. This criterion rewards complementary directions important to evaluation. With dependent trajectories, independent-frame counting is inappropriate; with misspecification, design-dependent bias must also be included \citep{white1982}. Our public-data study evaluates a specified matrix criterion rather than claiming that it equals a trained policy's risk.

\subsection{Representation and policy expressiveness}\label{sec:repr}
For a fixed expert joint law and deterministic $Z=g(H)$, any policy satisfies
\begin{equation}\label{eq:repr}
\E_H\KL(p(A\mid H)\|\pi(A\mid Z))
=I(A;H\mid Z)+\E_Z\KL(p(A\mid Z)\|\pi(A\mid Z)).
\end{equation}
Thus the best $N$-parameter policy has population approximation floor
\begin{equation}\label{eq:reprfloor}
\mathcal E_g(N)=I(A;H\mid Z)+\inf_{\pi\in\Pi_{g,N}}\E_Z\KL(p(A\mid Z)\|\pi(A\mid Z)).
\end{equation}
The first term measures action information discarded before prediction; the second concerns what the policy family can express. Increasing the head alone can address only the second. These are fixed-law population quantities, distinct from finite-data oracle gaps. Moreover, a lossy representation can merge histories with different actions. Changing their mixture then changes $p_q(A\mid Z)$ even if $p(A\mid H)$ is invariant, coupling data design and representation. Section~\ref{sec:numerical} gives a finite example.

\subsection{Optimization and generalization}\label{sec:optimization}
Let empirical NLL $\hat C_S$ be $L_s$-smooth, have finite infimum $\hat C_S^\star$, and satisfy $\|\nabla\hat C_S(\theta)\|^2/2\ge\mu(\hat C_S(\theta)-\hat C_S^\star)$ along full-batch gradient-descent iterates. For $0<\eta\le1/L_s$,
\begin{equation}\label{eq:pl}
\hat C_S(\theta_t)-\hat C_S^\star\le(1-\eta\mu)^t[\hat C_S(\theta_0)-\hat C_S^\star].
\end{equation}
This is an empirical convergence bound under a Polyak--\L{}ojasiewicz condition \citep{karimi2016}. Define the uniform discrepancy between empirical and reference NLL as $u_S$. An empirical $\varepsilon$-optimal output satisfies
\begin{align}
u_S&=\sup_{\pi\in\Pi}|\hat C_S(\pi)-C_E(\pi)|,\nonumber\\
\R_E(\pi_t)-\inf_{\pi\in\Pi}\R_E(\pi)&\le\varepsilon+2u_S.\label{eq:optgen}
\end{align}
Additional steps can reduce optimization error while sampling uncertainty or target mismatch remains. Distinct demonstrations, repeated presentations, stopping rules, and compute therefore require separate accounting.

\subsection{Deployment contexts}\label{sec:shift}
For a fixed policy and expert conditional, write $\ell_\pi(h)=\KL(p_E(\cdot\mid h)\|\pi(\cdot\mid h))$. A change of history measure gives
\begin{equation}\label{eq:shift}
\Delta_{\rm shift}(\pi)=\int[d_{\rm dep}(h)-d_E(h)]\ell_\pi(h)\,dh.
\end{equation}
When $d_{\rm dep}\ll d_E$, this is $\operatorname{Cov}_{d_E}(w,\ell_\pi)$ for $w=d_{\rm dep}/d_E$. It can have either sign: what matters is the alignment of deployment frequency with local imitation loss. Raw NLL shifts include a second contribution,
\begin{equation}\label{eq:shift-nll}
C_{\rm dep}(\pi)-C_E(\pi)=\Delta_{\rm shift}(\pi)
+\int[d_{\rm dep}(h)-d_E(h)]\mathsf H[p_E(\cdot\mid h)]\,dh.
\end{equation}
The entropy contribution can reverse the observed direction. This fixed-policy identity also differs from comparing policies whose actions induce different occupancies.

\section{Numerical checks}\label{sec:numerical}
\input{sections/numerical}

\section{Information design on public trajectories}\label{sec:open-data}
\input{sections/open_data}

\section{Discussion}
The five relations describe conditional interventions rather than independent additive mechanisms. Data allocation depends on the representation through its information matrix; changing the representation can alter the conditional target seen by the predictor; and converged training can leave a population gap. A practical comparison must specify what remains fixed at every intervention.

The numerical checks provide known-target tests of the analytical constructions. The public-trajectory study supplies a different kind of evidence: measured feature geometry can favor allocations that differ from evaluation frequencies, while balanced targets can yield no improvement. Its criterion is conditional on the chosen features, strata, and covariance model. It does not measure imitation loss, establish learned-policy Fisher information, or demonstrate task success. The distinction matters because a useful geometric design can still be undermined by model bias, representation aliasing, or optimization.

\paragraph{Scope and limitations.}
The theory applies under the assumptions attached to each relation. Finite-description bounds do not estimate unrestricted neural capacity; local Fisher approximations can fail under singularity, misspecification, or changing features. The data study concerns one public manipulation dataset and fixed state features, rather than language-conditioned or cross-embodiment generalization. Fixed-policy reweighting excludes changes in expert conditionals and policy-induced occupancy. These boundaries motivate future comparisons of predictive likelihood and closed-loop performance as separate outcomes.

\section{Conclusion}
A common prediction target turns broad scaling questions into specific resource-allocation comparisons. Capacity and data interact, useful demonstrations depend on evaluation priorities, representation loss differs from policy approximation, and empirical convergence differs from deployment accuracy. Analytical checks and public-trajectory matrix calculations illustrate these distinctions, including null results and an NLL--KL sign reversal. The resulting prescription is conditional: identify the bottleneck, state what is held fixed, and evaluate the corresponding intervention with the appropriate quantity.

\paragraph{Use of generative AI tools.}
OpenAI Codex assisted with manuscript drafting and revision, code development, and numerical analysis. Generative image tools assisted with the conceptual illustration in Figure~\ref{fig:framework}. The authors are responsible for the scientific claims, references, and results.

\label{end:main}
\clearpage
\bibliographystyle{plainnat}
\bibliography{references}
\clearpage
\appendix
\section{Proofs of the conditional relations}\label{app:proofs}
\subsection{Gaussian calibration}
For Gaussian distributions with common covariance $\sigma^2I$, the conditional prediction KL is $\|\theta-\hat\theta\|^2/(2\sigma^2)$. Given $D$ observations, coordinate $j$ has posterior variance and mean
\begin{equation}
 v_j=(\tau_j^{-2}+D/\sigma^2)^{-1},\qquad
 m_j=v_j(D/\sigma^2)\bar A_j.
\end{equation}
The posterior mean minimizes conditional squared loss in each trainable coordinate, giving Bayes squared error $v_j$. A coordinate fixed to zero contributes unconditional error $\tau_j^2$. Summing proves Eq.~\eqref{eq:exactgaussian}; at $D=0$, the posterior means equal zero. Among subsets of $N$ coordinates chosen before observing data, training a coordinate with prior variance $x$ reduces unscaled error by
\begin{equation}
 r(x)=\frac{Dx^2}{\sigma^2+Dx},\qquad
 r'(x)=\frac{Dx(2\sigma^2+Dx)}{(\sigma^2+Dx)^2}\ge0.
\end{equation}
Thus selecting the largest prior variances is optimal. For fixed $N$ and positive selected variances,
\begin{equation}
 L_G^\star(N,D)=\frac{\sum_{j>N}\tau_j^2}{2\sigma^2}+\frac{N}{2D}+O(D^{-2}).
\end{equation}
The expansion need not be uniform when the dimension or selected spectrum changes with $D$.

\subsection{Conditional compute allocation}
Substitute $D=F/(\kappa N)$ into the reducible terms of Eq.~\eqref{eq:candidate}. Differentiating with respect to $\log N$ gives
\begin{equation}
 -\alpha aN^{-\alpha}+\beta c(\kappa N/F)^\beta.
\end{equation}
Its derivative with respect to $\log N$ is strictly positive, so the unique positive stationary point is the minimum:
\begin{equation}\label{eq:compute}
 N^\star=\left(\frac{\alpha a}{\beta c}\right)^{1/(\alpha+\beta)}(F/\kappa)^{\beta/(\alpha+\beta)},\qquad
 D^\star=\left(\frac{\beta c}{\alpha a}\right)^{1/(\alpha+\beta)}(F/\kappa)^{\alpha/(\alpha+\beta)}.
\end{equation}
This interior formula is subject to the feasible budget range. Equalizing $aN^{-\alpha}$ and $cD^{-\beta}$ instead gives $D_\times(N)=(c/a)^{1/\beta}N^{\alpha/\beta}$.

\subsection{Local information design}
Assume the maximum-likelihood estimate is consistent, $\sqrt D(\hat\theta_D-\theta_0)\Rightarrow\mathcal N(0,F(q)^{-1})$, and its scaled second moments converge to $F(q)^{-1}$. Also assume the expected evaluation-risk Taylor remainder is $o(D^{-1})$. These are assumptions about the actual conditional model at its fixed input representation, including invariance across strata.

At the true parameter, risk and its gradient vanish. With $u=\hat\theta_D-\theta_0$, Taylor expansion yields $\R_E(p_{\hat\theta_D})=u^{\mathsf T}G_Eu/2+r_D$. Taking expectations and using second-moment convergence gives
\begin{equation}
 D\E[u^{\mathsf T}G_Eu]\longrightarrow\tr[G_EF(q)^{-1}],
\end{equation}
which proves Eq.~\eqref{eq:fisher}. Subtracting expansions proves the comparison for two fixed admissible designs. Asymptotic normality alone does not supply the needed moment or remainder control; optimizing a $D$-dependent design requires a uniform version.

The matrix derivative $\partial\mathcal I^{-1}/\partial n_i=-\mathcal I^{-1}F_i\mathcal I^{-1}$ proves Eq.~\eqref{eq:marginal}. Convexity follows from the second directional derivative
\begin{equation}
 \frac{d^2}{dt^2}\frac12\tr[G_E(\mathcal I+tB)^{-1}]\bigg|_{t=0}
 =\tr(G_E\mathcal I^{-1}B\mathcal I^{-1}B\mathcal I^{-1})\ge0.
\end{equation}
Indeed, the final expression is $\tr(CX^2)$, where
\[
 C=\mathcal I^{-1/2}G_E\mathcal I^{-1/2}\succeq0,
 \qquad X=\mathcal I^{-1/2}B\mathcal I^{-1/2}
\]
and $X$ is symmetric. The KKT conditions therefore equalize $v_i/c_i$ among active strata; inactive strata have no greater value. This concerns the continuous criterion, rather than a noisy finite-batch greedy implementation.

\subsection{Representation and mixture interaction}
Insert $p(A\mid Z)$ into the log-density ratio defining the left side of Eq.~\eqref{eq:repr}. The expectation of $\log[p(A\mid H)/p(A\mid Z)]$ is $I(A;H\mid Z)$ because $Z=g(H)$. The remaining ratio depends only on $(A,Z)$ and has expectation $\E_Z\KL(p(A\mid Z)\|\pi(A\mid Z))$. Taking the policy infimum affects only the second term, proving Eq.~\eqref{eq:reprfloor}. The information term vanishes precisely when $Z$ is action sufficient, up to null sets.

\begin{proposition}[Mixture-induced aliasing]\label{prop:aliasing}
Let $H=(Z,S)$ and suppose acquisition changes $p_q(S\mid Z)$ while preserving $p(A\mid Z,S)$. Then
\begin{equation}
 p_q(A\mid Z)=\sum_s p(A\mid Z,S=s)\,p_q(S=s\mid Z).
\end{equation}
Therefore invariance before discarding $S$ does not imply invariance at the retained input $Z$.
\end{proposition}
\begin{proof}
The equality is the law of total probability. If two discarded states have distinct conditional action laws and their conditional mixture weights change, the resulting convex combination can change. In the binary-state case it changes whenever the weight changes and the two laws differ.\end{proof}
This gives a direct reason to check conditional invariance after compression before applying a fixed-model information-design theorem.

\subsection{Convergence and population comparison}
The descent lemma with $\eta\le1/L_s$ gives
\begin{equation}
 \hat C_S(\theta_{t+1})\le\hat C_S(\theta_t)-\frac\eta2\|\nabla\hat C_S(\theta_t)\|^2
 \le\hat C_S(\theta_t)-\eta\mu[\hat C_S(\theta_t)-\hat C_S^\star].
\end{equation}
Subtract the infimum and iterate to obtain Eq.~\eqref{eq:pl}. For the population comparison, empirical $\varepsilon$-optimality and uniform deviation imply, for every $\pi\in\Pi$,
\begin{equation}
 C_E(\pi_t)\le\hat C_S(\pi_t)+u_S
 \le\hat C_S(\pi)+\varepsilon+u_S
 \le C_E(\pi)+\varepsilon+2u_S.
\end{equation}
Taking the infimum and subtracting the common conditional entropy proves Eq.~\eqref{eq:optgen}. A finite-data control of $u_S$ requires a separate complexity and sampling argument.

\subsection{Deployment reweighting}
Subtracting the two integrals defining reference and deployment risk proves Eq.~\eqref{eq:shift}. When $w=d_{\rm dep}/d_E$ exists, $\E_{d_E}w=1$ gives $\E_{d_E}[(w-1)\ell_\pi]=\operatorname{Cov}_{d_E}(w,\ell_\pi)$. Cauchy--Schwarz yields
\begin{equation}
 |\Delta_{\rm shift}(\pi)|\le\sqrt{\operatorname{Var}_{d_E}(w)\operatorname{Var}_{d_E}(\ell_\pi)}.
\end{equation}
For each history, expected log loss is the sum of expert conditional entropy and local KL. Integrating their sum against $d_{\rm dep}-d_E$ proves Eq.~\eqref{eq:shift-nll}. No entropy cancellation across different history laws is assumed.

\section{A finite-description capacity guarantee}\label{app:finite}
Fix a finite action alphabet and a codebook $\Pi_N$ of $K_N$ policies before observing the data. Assume $\pi(a\mid h)\ge\epsilon>0$, let $M=\log(1/\epsilon)$, and suppose $K_N\le2^{C_N}$. Here $C_N$ counts all policy-specific description bits; relating it to a raw parameter count is an additional coding assumption. For IID data from the fixed evaluation joint law, define
\begin{equation}
 a_N=\sup_\xi\min_{\pi\in\Pi_N}\R_\xi(\pi),\qquad
 \mathfrak R_N(D)=\inf_{\cA:S_D\mapsto\Pi_N}\sup_\xi\E\R_\xi(\cA(S_D)).
\end{equation}
This minimax statement is separate from the Bayes aggregation in Eq.~\eqref{eq:bayesfunctional}.
\begin{theorem}\label{thm:codebook}
Under these assumptions,
\begin{equation}\label{eq:finitebound}
 a_N\le\mathfrak R_N(D)\le a_N+2M\sqrt{\frac{\log(2K_N)}{2D}}
 \le a_N+2M\sqrt{\frac{C_N\log2+\log2}{2D}}.
\end{equation}
\end{theorem}
\begin{proof}
Each negative log loss lies in $[0,M]$. Hoeffding's lemma bounds each signed empirical deviation by a sub-Gaussian moment generating function with variance proxy $M^2/(4D)$ \citep{hoeffding1963}. Bounding the maximum over $2K_N$ signed deviations yields
\begin{equation}
 \E\sup_{\pi\in\Pi_N}|\hat C_S(\pi)-C(\pi)|\le M\sqrt{\frac{\log(2K_N)}{2D}}.
\end{equation}
The empirical-minimizer comparison used above gives excess population cross-entropy at most twice this deviation; expert entropy cancels on the common target. Taking the supremum over experts proves the upper bound. Every output, including a randomized output, has risk at least the best codeword for that expert. Taking the supremum and then the algorithm infimum proves the lower bound.\end{proof}
The bound separates an approximation curve $a_N$ from a sufficient estimation term without specifying a power-law approximation exponent.

\input{sections/numerical_details}
\end{document}

%% file: sections/numerical.tex
\subsection{Analytical calibration and counterexamples}\label{subsec:numerical-calibration}
We evaluate specified distributions to distinguish the identities' computational consequences from empirical claims about robot policies.
In the Gaussian model, $m=16$, $\sigma^2=1$, and $\tau_j^2=j^{-1.3}$. Across $N\in\{2,8,16\}$ and $D\in\{4,16,64,256\}$, $100{,}000$ independent draws per setting agree with the exact Bayes risk to within 0.14\% relative deviation. The largest discrepancy is $1.25$ estimated Monte Carlo standard errors.

A four-context Bernoulli example gives an information-loss term of $0.05123$ nats after merging contexts in pairs. Changing the sampling mixture changes the optimal action conditional at that coarsened input, adding a persistent mismatch even though the full-context expert is unchanged. A second calculation holds the policy fixed: target A increases KL by $0.04200$ nats while decreasing NLL by $0.06495$ nats. The $-0.10695$-nat conditional-entropy change explains the opposite signs. Thus cross-distribution NLL differences cannot generally be read as imitation-KL shifts.

Finally, exact trajectories of a positive-definite quadratic satisfy the stated PL bound for the two admissible step sizes; a step above $2/L_s$ diverges. These checks illustrate the roles of the assumptions and verify the calculations, rather than validate a neural-policy convergence law.

\begin{table}[!htbp]\centering\small
\begin{tabular}{lrrr}\toprule
Target & $\Delta\mathrm{KL}$ & $\Delta\mathsf H$ & $\Delta\mathrm{NLL}$\\\midrule
A & $+0.04200$ & $-0.10695$ & $-0.06495$\\
B & $-0.04297$ & $+0.04410$ & $+0.00113$\\\bottomrule
\end{tabular}
\caption{Exact finite-distribution changes in nats for a fixed policy. Conditional-entropy changes account for the different KL and NLL signs.}\label{tab:numerical-checks}
\end{table}

%% file: sections/open_data.tex
\subsection{Protocol and measured quantity}
We examine how public demonstration-state geometry changes an information-design prescription without fitting an action predictor. The LeRobot PushT release contains 206 simulated pushing demonstrations and 25,650 state--action records at 10 Hz \citep{pushtdata,chi2023diffusion}. We use the fixed dataset revision \texttt{7628202a2180}; the full commit and file hashes accompany the code. A fixed episode permutation assigns 144 episodes (17,858 records) to design calibration and 62 episodes (7,792 records) to evaluation. No episode appears in both sets.

We divide the pusher positions at horizontal coordinate 256 into left and right strata. We normalize each position coordinate by dividing by 256 and subtracting 1. The primary fixed feature map is $\phi=(1,x,y,xy,x^2,y^2)$. We also evaluate $\phi=(1,x,y)$ as a prespecified sensitivity check. For each split $s$ and stratum $i$, let $\widehat F_{s,i}$ be the mean of $\phi\phi^{\mathsf T}$ over its recorded states. These matrices would be Fisher information for a scalar unit-variance linear Gaussian reference model. They are \emph{not} estimated Fisher matrices of a trained robot policy.

For left-target weights $w\in\{0.2,0.5,0.8\}$, we evaluate
\begin{equation}\label{eq:publicdesign}
\widehat V_s(q;w)=\tr\!\left[\bigl(w\widehat F_{s,L}+(1-w)\widehat F_{s,R}\bigr)
\bigl(q\widehat F_{s,L}+(1-q)\widehat F_{s,R}+10^{-6}I\bigr)^{-1}\right].
\end{equation}
The ridge is a declared numerical regularizer, so this is a regularized design criterion. We choose $\hat q$ on the calibration split from $\{0.020,0.025,\ldots,0.980\}$, then hold it fixed when recomputing the criterion from evaluation states. Comparators are uniform allocation $q=0.5$, target matching $q=w$, and the calibration frame proportion $q=0.5736$. Feature maps, strata, target weights, grid, and split seed were fixed before inspecting the results. All six cases are reported. Uncertainty is assessed with 1,000 paired resamples of whole evaluation episodes, keeping the calibrated allocation fixed; percentile intervals describe held-out moment variability conditional on this calibration split.

\begin{table}[!htbp]
\centering\small
\setlength{\tabcolsep}{4.2pt}
\begin{tabular}{llrrrrl}
\toprule
Basis & $w$ & $\hat q$ & $V(\hat q)$ & $V(0.5)$ & $V(w)$ & Reduction [95\% interval] \\
\midrule
Linear & 0.2 & 0.350 & 2.864 & 3.008 & 3.000 & 4.52 [3.64, 5.55] \\
       & 0.5 & 0.510 & 3.001 & 3.000 & 3.000 & $-0.03$ [$-0.09$, 0.04] \\
       & 0.8 & 0.670 & 2.857 & 2.992 & 3.000 & 4.77 [3.87, 5.65] \\
Quadratic & 0.2 & 0.335 & 5.704 & 6.086 & 6.000 & 4.93 [4.10, 5.97] \\
          & 0.5 & 0.500 & 6.000 & 6.000 & 6.000 & 0.00 [0.00, 0.00] \\
          & 0.8 & 0.660 & 5.626 & 5.914 & 6.000 & 6.23 [5.13, 7.17] \\
\bottomrule
\end{tabular}
\caption{Held-out design criteria and relative reductions $100[1-V(\hat q)/V(w)]$ (\%). Intervals use paired whole-episode resampling. Positive values favor the calibrated allocation; the quantities describe design geometry rather than policy performance.}\label{tab:publicdesign}
\end{table}

\subsection{Findings and limits}
For unequal target weights, the calibrated allocations lie between uniform and target-matched sampling. On the primary quadratic basis they lower the held-out criterion by 4.93\% and 6.23\% relative to target matching (Table~\ref{tab:publicdesign}). Equal target weights yield no reduction for the quadratic basis and a small negative transfer for the linear basis ($-0.03\%$, interval spanning zero). Thus adapting allocation is useful in these asymmetric cases, but is not an automatic improvement. The two feature maps give different prescriptions, consistent with information design being conditional on the representation.

This analysis uses existing simulated demonstration states and never fits an action distribution. It neither measures expert KL nor tests action success, a neural scaling exponent, or the asymptotic $1/D$ law. Adjacent frames are correlated; they are not counted as independent demonstrations. The within-stratum moments are frame-weighted, while the uncertainty resampling unit is the episode. Finally, the hypothetical allocations reweight state strata; physically acquiring those proportions in complete robot trajectories may have additional costs or feasibility constraints.

%% file: sections/numerical_details.tex
\section{Numerical specifications}\label{app:numerical-details}
The Gaussian calculation uses independent $\theta_j\sim\mathcal N(0,j^{-1.3})$ and $\bar A_j=\theta_j+\varepsilon_j/\sqrt D$ with independent standard-normal $\varepsilon_j$. The posterior mean is $\hat\theta_j=D\tau_j^2\bar A_j/(1+D\tau_j^2)$ for $j\le N$ and zero otherwise. We average $\|\hat\theta-\theta\|^2/2$ over $100{,}000$ draws. Common random numbers pair the settings; each setting's draws remain independent. The random seed is $20261112$.

The finite example has reference context probabilities $(0.1,0.2,0.3,0.4)$ and Bernoulli expert probabilities $(0.05,0.2,0.6,0.9)$. The representation merges contexts $1,2$ and $3,4$. The two training mixtures are $(0.25,0.05,0.1,0.6)$ and $(0.05,0.5,0.4,0.05)$; their represented action conditionals are computed by exact weighted averaging. For the shift calculation, the fixed policy predicts $(0.2,0.2,0.7,0.7)$, and target mixtures A and B are $(0.05,0.05,0.05,0.85)$ and $(0.1,0.65,0.2,0.05)$. All KL and entropy quantities are exact finite sums in nats.

The quadratic is $f(x)=x^\mathsf T\operatorname{diag}(1,4,12)x/2$, with $x_0=(2,-1,0.5)$. Thus $\mu=1$ and $L_s=12$. The objective values are evaluated directly as
\begin{equation}
f_t=\frac12\sum_j\lambda_jx_{0j}^2(1-\eta\lambda_j)^{2t},
\qquad 0\le t\le60,
\end{equation}
for $\eta\in\{1/24,1/12,0.18\}$. This is a closed-form evaluation of the quadratic recurrence.

%% file: main.bbl
\begin{thebibliography}{18}
\providecommand{\natexlab}[1]{#1}
\providecommand{\url}[1]{\texttt{#1}}
\expandafter\ifx\csname urlstyle\endcsname\relax
  \providecommand{\doi}[1]{doi: #1}\else
  \providecommand{\doi}{doi: \begingroup \urlstyle{rm}\Url}\fi

\bibitem[Black et~al.(2024)Black, Brown, Driess, Esmail, Equi, Finn, Fusai,
  Groom, Hausman, Ichter, Jakubczak, Jones, Ke, Levine, Li-Bell, Mothukuri,
  Nair, Pertsch, Shi, Tanner, Vuong, Walling, Wang, and Zhilinsky]{black2024}
Kevin Black, Noah Brown, Danny Driess, Adnan Esmail, Michael Equi, Chelsea
  Finn, Niccolo Fusai, Lachy Groom, Karol Hausman, Brian Ichter, Szymon
  Jakubczak, Tim Jones, Liyiming Ke, Sergey Levine, Adrian Li-Bell, Mohith
  Mothukuri, Suraj Nair, Karl Pertsch, Lucy~Xiaoyang Shi, James Tanner, Quan
  Vuong, Anna Walling, Haohuan Wang, and Ury Zhilinsky.
\newblock {$\pi_0$}: A vision-language-action flow model for general robot
  control, 2024.
\newblock URL \url{https://arxiv.org/abs/2410.24164}.

\bibitem[Brohan et~al.(2023)Brohan, Brown, Carbajal, Chebotar, Chen,
  Choromanski, Ding, Driess, Dubey, Finn, et~al.]{brohan2023}
Anthony Brohan, Noah Brown, Justice Carbajal, Yevgen Chebotar, Xi~Chen,
  Krzysztof Choromanski, Tianli Ding, Danny Driess, Avinava Dubey, Chelsea
  Finn, et~al.
\newblock {RT-2}: Vision-language-action models transfer web knowledge to
  robotic control, 2023.
\newblock URL \url{https://arxiv.org/abs/2307.15818}.

\bibitem[Chi et~al.(2023)Chi, Feng, Du, Xu, Cousineau, Burchfiel, and
  Song]{chi2023diffusion}
Cheng Chi, Siyuan Feng, Yilun Du, Zhenjia Xu, Eric Cousineau, Benjamin
  Burchfiel, and Shuran Song.
\newblock Diffusion policy: Visuomotor policy learning via action diffusion.
\newblock In \emph{Robotics: Science and Systems}, 2023.
\newblock URL \url{https://arxiv.org/abs/2303.04137}.

\bibitem[Cover and Thomas(2006)]{cover2006}
Thomas~M. Cover and Joy~A. Thomas.
\newblock \emph{Elements of Information Theory}.
\newblock Wiley, 2 edition, 2006.
\newblock \doi{10.1002/047174882X}.
\newblock URL \url{https://doi.org/10.1002/047174882X}.

\bibitem[Gneiting and Raftery(2007)]{gneiting2007}
Tilmann Gneiting and Adrian~E. Raftery.
\newblock Strictly proper scoring rules, prediction, and estimation.
\newblock \emph{Journal of the American Statistical Association}, 102\penalty0
  (477):\penalty0 359--378, 2007.
\newblock \doi{10.1198/016214506000001437}.
\newblock URL \url{https://doi.org/10.1198/016214506000001437}.

\bibitem[Hejna et~al.(2025)Hejna, Bhateja, Jiang, Pertsch, and
  Sadigh]{hejna2025}
Joey Hejna, Chethan~Anand Bhateja, Yichen Jiang, Karl Pertsch, and Dorsa
  Sadigh.
\newblock {ReMix}: Optimizing data mixtures for large scale imitation learning.
\newblock In \emph{Proceedings of The 8th Conference on Robot Learning}, volume
  270 of \emph{Proceedings of Machine Learning Research}, pages 145--164. PMLR,
  2025.
\newblock URL \url{https://proceedings.mlr.press/v270/hejna25a.html}.

\bibitem[Hoeffding(1963)]{hoeffding1963}
Wassily Hoeffding.
\newblock Probability inequalities for sums of bounded random variables.
\newblock \emph{Journal of the American Statistical Association}, 58\penalty0
  (301):\penalty0 13--30, 1963.
\newblock \doi{10.1080/01621459.1963.10500830}.
\newblock URL \url{https://doi.org/10.1080/01621459.1963.10500830}.

\bibitem[Hoffmann et~al.(2022)Hoffmann, Borgeaud, Mensch, Buchatskaya, Cai,
  Rutherford, de~Las~Casas, Hendricks, Welbl, Clark, Hennigan, Noland,
  Millican, van~den Driessche, Damoc, Guy, Osindero, Simonyan, Elsen, Rae,
  Vinyals, and Sifre]{hoffmann2022}
Jordan Hoffmann, Sebastian Borgeaud, Arthur Mensch, Elena Buchatskaya, Trevor
  Cai, Eliza Rutherford, Diego de~Las~Casas, Lisa~Anne Hendricks, Johannes
  Welbl, Aidan Clark, Tom Hennigan, Eric Noland, Katie Millican, George van~den
  Driessche, Bogdan Damoc, Aurelia Guy, Simon Osindero, Karen Simonyan, Erich
  Elsen, Jack~W. Rae, Oriol Vinyals, and Laurent Sifre.
\newblock Training compute-optimal large language models, 2022.
\newblock URL \url{https://arxiv.org/abs/2203.15556}.

\bibitem[Kaplan et~al.(2020)Kaplan, McCandlish, Henighan, Brown, Chess, Child,
  Gray, Radford, Wu, and Amodei]{kaplan2020}
Jared Kaplan, Sam McCandlish, Tom Henighan, Tom~B. Brown, Benjamin Chess, Rewon
  Child, Scott Gray, Alec Radford, Jeffrey Wu, and Dario Amodei.
\newblock Scaling laws for neural language models, 2020.
\newblock URL \url{https://arxiv.org/abs/2001.08361}.

\bibitem[Karimi et~al.(2016)Karimi, Nutini, and Schmidt]{karimi2016}
Hamed Karimi, Julie Nutini, and Mark Schmidt.
\newblock Linear convergence of gradient and proximal-gradient methods under
  the {Polyak--{\L}ojasiewicz} condition, 2016.
\newblock URL \url{https://arxiv.org/abs/1608.04636}.

\bibitem[Kim et~al.(2024)Kim, Pertsch, Karamcheti, Xiao, Balakrishna, Nair,
  Rafailov, Foster, Lam, Sanketi, Vuong, Kollar, Burchfiel, Tedrake, Sadigh,
  Levine, Liang, and Finn]{kim2024}
Moo~Jin Kim, Karl Pertsch, Siddharth Karamcheti, Ted Xiao, Ashwin Balakrishna,
  Suraj Nair, Rafael Rafailov, Ethan Foster, Grace Lam, Pannag Sanketi, Quan
  Vuong, Thomas Kollar, Benjamin Burchfiel, Russ Tedrake, Dorsa Sadigh, Sergey
  Levine, Percy Liang, and Chelsea Finn.
\newblock {OpenVLA}: An open-source vision-language-action model, 2024.
\newblock URL \url{https://arxiv.org/abs/2406.09246}.

\bibitem[{LeRobot Contributors}(2024)]{pushtdata}
{LeRobot Contributors}.
\newblock {PushT} demonstration dataset.
\newblock Hugging Face Datasets, 2024.
\newblock URL \url{https://huggingface.co/datasets/lerobot/pusht}.
\newblock Revision 7628202a2180; 206 episodes, 25,650 frames.

\bibitem[Lin et~al.(2024)Lin, Hu, Sheng, Wen, You, and Gao]{lin2024scaling}
Fanqi Lin, Yingdong Hu, Pingyue Sheng, Chuan Wen, Jiacheng You, and Yang Gao.
\newblock Data scaling laws in imitation learning for robotic manipulation,
  2024.
\newblock URL \url{https://arxiv.org/abs/2410.18647}.

\bibitem[Pearce et~al.(2025)Pearce, Rashid, Bignell, Georgescu, Devlin, and
  Hofmann]{pearce2025}
Tim Pearce, Tabish Rashid, David Bignell, Raluca Georgescu, Sam Devlin, and
  Katja Hofmann.
\newblock Scaling laws for pre-training agents and world models.
\newblock In \emph{Proceedings of the 42nd International Conference on Machine
  Learning}, volume 267 of \emph{Proceedings of Machine Learning Research},
  pages 48542--48562. PMLR, 2025.
\newblock URL \url{https://proceedings.mlr.press/v267/pearce25a.html}.

\bibitem[Pukelsheim(2006)]{pukelsheim2006}
Friedrich Pukelsheim.
\newblock \emph{Optimal Design of Experiments}, volume~50 of \emph{Classics in
  Applied Mathematics}.
\newblock Society for Industrial and Applied Mathematics, 2006.
\newblock \doi{10.1137/1.9780898719109}.
\newblock URL \url{https://epubs.siam.org/doi/book/10.1137/1.9780898719109}.

\bibitem[Ross et~al.(2011)Ross, Gordon, and Bagnell]{ross2011}
Stephane Ross, Geoffrey Gordon, and Drew Bagnell.
\newblock A reduction of imitation learning and structured prediction to
  no-regret online learning.
\newblock In \emph{Proceedings of the Fourteenth International Conference on
  Artificial Intelligence and Statistics}, volume~15 of \emph{Proceedings of
  Machine Learning Research}, pages 627--635. PMLR, 2011.
\newblock URL \url{https://proceedings.mlr.press/v15/ross11a.html}.

\bibitem[van~der Vaart(1998)]{vaart1998}
A.~W. van~der Vaart.
\newblock \emph{Asymptotic Statistics}, volume~3 of \emph{Cambridge Series in
  Statistical and Probabilistic Mathematics}.
\newblock Cambridge University Press, 1998.
\newblock \doi{10.1017/CBO9780511802256}.
\newblock URL \url{https://doi.org/10.1017/CBO9780511802256}.

\bibitem[White(1982)]{white1982}
Halbert White.
\newblock Maximum likelihood estimation of misspecified models.
\newblock \emph{Econometrica}, 50\penalty0 (1):\penalty0 1--25, 1982.
\newblock URL \url{https://www.jstor.org/stable/1912526}.

\end{thebibliography}
